\documentclass[final]{cvpr}

\usepackage[pagebackref=true,breaklinks=true,letterpaper=true,colorlinks,bookmarks=false]{hyperref}

\usepackage{times}
\usepackage{epsfig}
\usepackage{graphicx}
\usepackage{amsmath}
\usepackage{amssymb}
\usepackage{bbm}
\usepackage{comment}
\usepackage[ruled,lined]{algorithm2e}
\usepackage{makecell}
\usepackage{multirow}
\usepackage{stmaryrd}
\usepackage{subcaption}
\usepackage{float}
\usepackage{amsmath,amsfonts,bm}
\usepackage{cleveref}

\DeclareMathAlphabet{\mathsfit}{\encodingdefault}{\sfdefault}{m}{sl}
\SetMathAlphabet{\mathsfit}{bold}{\encodingdefault}{\sfdefault}{bx}{n}

\newcommand{\junwei}[1]{{\color{green}{(Junwei\@: #1)}}}
\newcommand{\llcao}[1]{{\color{blue}{(llcao\@: #1)}}}

\newcommand{\eat}[1]{} 

\newcommand{\fancyname}{\textit{STAN}}

\def\cvprPaperID{11407} 
\def\confYear{CVPR 2021}

\begin{document}

\title{
Spatial-Temporal Alignment Network for Action Recognition and Detection
}

\author{
Junwei Liang\textsuperscript{1}\thanks{Work partially done during a research internship at Google.} \qquad
Liangliang Cao\textsuperscript{2} \qquad
Xuehan Xiong\textsuperscript{2} \qquad
Ting Yu\textsuperscript{2} \qquad
Alexander Hauptmann\textsuperscript{1} \\
\textsuperscript{1}Carnegie Mellon University  \qquad\qquad
\textsuperscript{2}Google Cloud AI\\
{\tt\small \{junweil,alex\}@cs.cmu.edu, \{llcao,xxman,yuti\}@google.com} 
}

\maketitle

\setlength{\abovedisplayskip}{2pt} \setlength{\belowdisplayskip}{3pt}

\begin{abstract}
This paper studies
how to introduce viewpoint-invariant feature representations that can help action recognition and detection. Although we have witnessed great progress of action recognition in the past decade, 
it remains challenging yet interesting how to efficiently model the geometric variations in large scale datasets. This paper proposes a novel Spatial-Temporal Alignment Network (\fancyname) that aims to learn geometric invariant representations for action recognition and action detection. The {\fancyname} model is very light-weighted and generic,
which could be plugged into existing action recognition
models like ResNet3D and the SlowFast with a very low extra computational cost.
We test our {\fancyname} model extensively on AVA, Kinetics-400, AVA-Kinetics, Charades, and Charades-Ego datasets. The experimental results show that the {\fancyname} model can consistently improve the state of the arts in both action detection and action recognition tasks.
\eat{
This paper  studies the problem of viewpoint-invariant feature representations that aim to have better generalization abilities for action recognition and detection.
We make two main contributions.
The first contribution is a new spatial-temporal alignment model.
Our model achieves significant improvement with low extra computation.
The second contribution is that our model achieves state-of-the-art performance on several datasets.
We refer to our model as {\fancyname}. 
}
We will release our data, models and code.
\end{abstract}

\section{Introduction}

Human vision can recognize video actions efficiently despite the variations of viewpoints. 
Convolutional neural networks (CNNs) \cite{cnn-lecun-eccv10,cnn-dtran-iccv15,cnn-Carreira-Zisserman-cvpr17,DBLP:journals/corr/SigurdssonDFG16,DBLP:conf/nips/FeichtenhoferPW16} 
fully utilize the power of GPUs/TPUs and employ spatial-temporal filters to recognize actions, which
outperforms traditional models including oriented filtering in space time (HOG3D) \cite{DBLP:conf/bmvc/KlaserMS08}, spatial-temporal interest points \cite{DBLP:conf/cvpr/LaptevMSR08}, motion history images \cite{DBLP:journals/tsmc/TianCLZ12}, and trajectories \cite{DBLP:conf/iccv/WangS13a}.
However, due to the high variations in space-time, the state of the art of action recognition is still far from satisfactory, compared with the success of 2D CNNs in image recognition \cite{he2016deep}.

A key challenge of action recognition is to capture the variations across space and time.
Since CNN assumes the filters share weights at different locations, it can not explicitly model the viewpoint changes and other variations. To solve such limitations, a practical way is to expand feature representations to accommodate a higher degree of freedom. 
For example, the two-stream network \cite{twostream-Simonyan-nips14} proposes to integrate optical flow with RGB features. More recently, SlowFast
\cite{feichtenhofer2019slowfast} combines both slow and fast pathways to learn different temporal information, and obtain good performance. However, such feature expanding approaches quickly lead to
cumbersome, high-dimensional feature maps, which not only make the computation more expensive but also miss the geometric interpretation of the subjects. 


\begin{figure}[t!]
	\centering
		\includegraphics[width=0.47\textwidth]{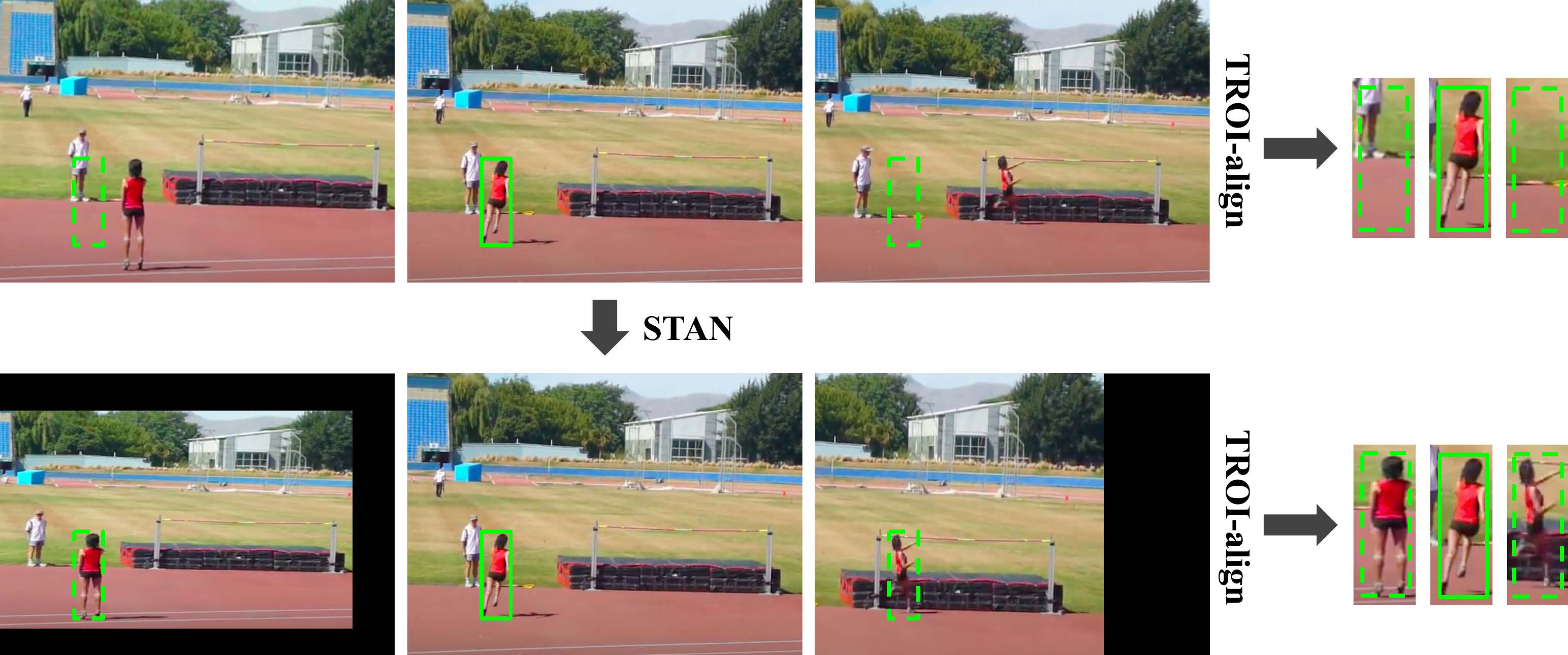} 
	\caption{An illustrative example of how {\fancyname} can address the bounding box misalignment issue in spatial temporal localization. In prior work~\cite{gu2018ava,feichtenhofer2019slowfast}, the detected person box (solid line) is expanded along the temporal axis (dotted line) to form a 3D cube for the subsequent temporal ROI-align (TROI-align). If the actor's movement is large across frames as shown in the top row (or there are substantial camera movements), the TROI-aligned features will be noisy with the majority being background pixels. {\fancyname} learns a spatial temporal transformation 
	that puts feature maps in the same coordinate system and in the bottom row example it is the central frame.}
	\label{fig:title}
\end{figure}

This paper proposes a different approach to capture the variation in action recognition. 
Instead of relying on stacking deeper CNN layers, this paper aims to explicitly learn geometric transformations and viewpoint invariant features. 
Our idea is motivated by \cite{kosiorek2019stacked}, which believes that  human vision relies on coordinate frames. 
However, the stacked capsule autoencoder in \cite{kosiorek2019stacked} is designed for 2D images, and too expensive for large scale visual recognition.

Fig.~\ref{fig:title} illustrates the importance of alignment in the problem of spatial temporal localization. Following the existing action detection pipeline~\cite{gu2018ava,feichtenhofer2019slowfast}, the action representation is obtained by first cropping a 3D cube within the spatial-temporal feature maps centered at the detected actor’s bounding box, followed by a Max Pooling operation. Without alignment, the representation is contaminated by background pixels as shown in the top row in Fig.~\ref{fig:title}. With alignment the representation as shown in the bottom row is much more focused on the actor.

In this paper,
we propose a Spatial Temporal Alignment Network ({\fancyname}) that aims to learn viewpoint invariant representations for action recognition and action detection. 
The {\fancyname} block is very light-weighted and generic, which could be plugged into existing action recognition models like ResNet3D, Non-Local Network~\cite{wang2018non} and the SlowFast network~\cite{feichtenhofer2019slowfast}. 
As discussed in Section~\ref{sec:arch}, we insert a {\fancyname} block between $res_2$ and $res_3$ in the ResNet3D architecture, 
and add it at the same location on the Fast pathway in the SlowFast model.
For the SlowFast + {\fancyname} model, the FLOPS increase is only \textbf{relatively 2.1\%} (134.5 GFLOPS vs. 131.7 GFLOPS) for action recognition on Charades, but we achieve \textbf{5\% relative} improvement on mean average precision.

The contribution of this paper is three-fold: 
(1) To the best of our knowledge, this is the first work to explore explicit spatial-temporal alignment in 3D CNNs for action detection.
(2) Our {\fancyname} requires very low extra FLOPS in addition to the backbone network. (3) Extensive experiments on different datasets suggest the model using {\fancyname} can outperform the state of the art.

\section{Related Work}


The research of action recognition has advanced with both new datasets and new models.  As one of the earliest action recognition benchmarks, KTH\cite{kth-icpr04} collects videos of individual actors repetitively performing six types of human actions (walking, jogging, running, boxing, hand waving and clapping) with a clean background. Because these videos are very simple, KTH dataset turns out to be a very easy benchmark since studies quickly obtained near-perfect accuracy on it \cite{DBLP:conf/cvpr/CaoLH10,DBLP:conf/cvpr/WangKSL11}. 

To overcome the limitation of KTH, 
the HMDB dataset \cite{hmdb51} was proposed in 2011 with 51 actions in 7000 video clips, while 
UCF101 \cite{ucf101} extended this effort by collecting 101 action classes in 13000 clips. Both benchmarks are captured 
with more diversified backgrounds. In the past decade, we have witnessed a steady improvement of accuracy on these two datasets by different methods including features fusion \cite{DBLP:journals/cviu/PengWWQ16}, two-stream network \cite{twostream-Simonyan-nips14}, C3D \cite{cnn-dtran-iccv15},  I3D \cite{cnn-Carreira-Zisserman-cvpr17}, graph-based approaches \cite{wang2018videos,chen2019graph,qi2018learning,zhang2019structured} and others \cite{xu2017r,chao2018rethinking,hou2017tube}. However, some clips in the UCF101 test set are taken from the same YouTube video as the training set \cite{kay2017kinetics}, which makes it relatively easy to obtain good accuracy on UCF101. As a result, the SOTA on UCF101 dataset is more than 98\%\cite{kay2017kinetics}. 

The modern benchmarks for action recognition and detection is the Kinetics dataset \cite{kay2017kinetics} and the AVA dataset \cite{li2020ava}, respectively. The Kinetics dataset proposes a bigger benchmark with more categories and more videos (e.g., 400 categories 160,000 clips in  \cite{kay2017kinetics}) as a harder benchmark. The action labels in AVA \cite{li2020ava} are annotated with spatial temporal locations, which is more challenging than the setting of one label per clip.
\eat{\llcao{shall we mention this or ignore this?}.
However, Kinetics and AVA datasets do not exhaust all the possible actions in all possible scales, for example, surveillance actions are not missing in the two datasets It is very likely that a even bigger dataset will appear in the future. 
}
Many new approaches ~\cite{tran2018closer,zhao2018trajectory,lin2019tsm,feichtenhofer2019slowfast,yang2020temporal} have been carried on these two datasets, of which the SlowFast network \cite{feichtenhofer2019slowfast} obtains the state of the art performance. Note that the SlowFast contains more parameters than C3D or I3D networks, by integrating features at both high and low frame rates. We can see the trend of action recognition in the last two decades is to collect bigger datasets (e.g. Kinetics) as well as build bigger models (e.g., I3D and SlowFast).

This paper considers action recognition using the spatial-temporal alignment to overcome the viewpoint variations in videos. In recent years, there has been a consistent effort to use alignment for image recognition
\cite{DBLP:conf/nips/HuangMLL12,Xiong_2013_CVPR,jaderberg2015spatial,lin2017inverse,kosiorek2019stacked}. 
Some data augmentation methods~\cite{liang2020garden,liang2020simaug} using 3D simulation have been proposed to tackle the viewpoint changes.
However, many previous studies show that alignment models are not as competitive as data-driven approaches like data augmentation or spatial pooling for image recognition.
Some recent works have to rely on very expensive models such as recurrent networks \cite{lin2017inverse} or stacked capsules \cite{kosiorek2019stacked}.
As a result, a lot of alignment-based recognition methods are limited to MNIST \cite{kosiorek2019stacked} and face recognition \cite{Xiong_2013_CVPR}. 
Some follow-up works on capsule network~\cite{duarte2018videocapsulenet} and 2D alignment network~\cite{huang2019part} have been proposed but they are limited to action recognition on small datasets like UCF101~\cite{ucf101} and JHMDB~\cite{jhuang2013towards}.
This paper shows that it is possible to build an efficient spatial-temporal alignment for both action recognition and detection, and improve the state of the arts with very few extra parameters.

\begin{figure*}[ht]
	\centering
		\includegraphics[width=0.95\textwidth]{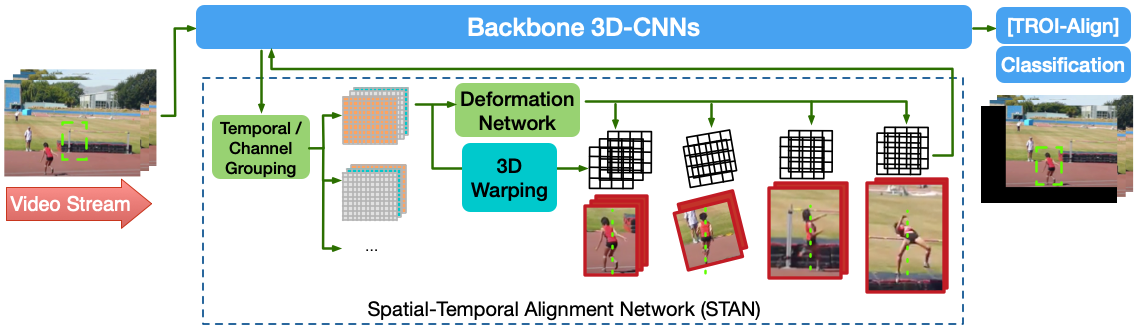} 
	\caption{Spatial Temporal Alignment Network ({\fancyname}) design. In our design, different temporal slices of the feature map can undergo different transformations to account for intra-clip camera motion and actor movement. The same applies to subgroups of the channel, an idea similar to multihead attention~\cite{vaswani2017attention}. We show four alignment examples with a reference line where the actor is aligned on for demonstration. After the features are transformed and processed through the backbone network, they are passed to a temporal ROI-Align layer (TROI-Align) if the task is action detection, and then finally to the classification layer.
	\eat{\llcao{ why is the most right box is called classification? seems more like a detection}\junwei{for detection, after TROI-align, it is also fed to a linear classification layer. I have added a alternative "TROI" module}.
	\llcao{seems  there is  hardly any relationship between the four photos and input? }\junwei{I guess the idea is to show that the person is somewhat ``align'' in the center, I have added a reference line in the frames}}
	}
	\label{fig:overview}
\end{figure*}

\section{The {\fancyname} Model}
\label{sec:method}

In this section, we describe our spatial-temporal alignment network for action recognition and detection, which we call {\fancyname}.
Motivated by the previous  works in image understanding \cite{DBLP:conf/nips/HuangMLL12,jaderberg2015spatial,lin2017inverse}, our work considers a generalized model in video domain, so that it can handle dynamic viewpoint changes in action recognition and detection.
The key idea of {\fancyname} is to utilize spatial-temporal alignment for feature maps to account for viewpoint changes and actor movements in the videos.
In general, given an input spatial-temporal feature map $\mathcal{I}_{in} \in \mathbf{R}^{C \times T \times H \times W}$ where $H$ stands for height, $W$ for width, $T$ for time and $C$ for channels, the alignment function is defined as
\begin{align}
\label{eqn:stan}
    \text{STAN}(\mathcal{I}_{in}) = \mathcal{T}(\theta, \mathcal{I}_{in}) + \mathcal{I}_{in},
    \text{where} \; \theta = \mathcal{D}(\mathcal{I}_{in})
\end{align}
\eat{The output feature map $\mathcal{I}_{out}$ has the same number of channel as the input, but could have different sizes in the temporal and spatial dimensions.
In this paper we only explore same-size transformation.
}
In this paper, the output of {\fancyname} function is of the same dimension as $\mathcal{I}_{in}$. 
Function $ \mathcal{D}$ represents the deformation network, where the feature map alignment parameter $\theta \in \mathbf{R}^{4 \times 4}$ is computed based on the input feature map.
Function $ \mathcal{D}$ can be in the form of a simple feed-forward network using spatial-temporal features \cite{tran2018closer}.
Function $ \mathcal{T}$ is defined as the warping function, where input feature maps are warped based on the alignment parameter.
In this paper, we add a residual connection between the input feature maps and output feature maps for faster training and avoiding boundary effect described in ~\cite{lin2017inverse}.
The {\fancyname} layer can be added to different locations of the backbone to account for the alignment needs for different level of feature maps.

\subsection{Network Architecture}
\noindent The overall {\fancyname} architecture is shown in Fig.~\ref{fig:overview}.
Our model uses a 3D CNN as the backbone network to extract spatial-temporal feature maps from video frames. In addition,  {\fancyname}
has the following key components:


\noindent\textbf{Deformation Network} produces the alignment / deformation parameter $\theta$ in Eqn.~\ref{eqn:stan}. 

\noindent\textbf{Warping Module} samples from the input feature maps based on $\theta$ and outputs the final transformed feature maps.

\noindent\textbf{Channel Grouping} allows different alignment for different subsets of the features similar to multihead attention~\cite{vaswani2017attention}.

\noindent\textbf{Temporal Grouping} learns a separate alignment for different temporal segments within the video clip.

In the following, we will introduce the above modules in details.

\subsection{Deformation Network}\label{sec:deformnet}
The deformation network $\mathcal{D}$ produces a transformation parameter, $\theta \in \mathbf{R}^{4 \times 4}$.
Our network is based on R(2+1)D~\cite{tran2018closer} although other options are possible, such as a simple feed-forward network, or a recurrent network and compositional function as proposed in ~\cite{lin2017inverse}.
Suppose the number of input channels is $C$, the details of our network architecture is presented in Table~\ref{tab:deformnet}.  
\begin{table}
\centering
\begin{tabular}{c | c | c }
Layer  & Filter size &  Output channels \\
\hline
conv & $1\times7^2$ & $C \sslash 4$ \\
\hline
conv & $7\times1^2$ & $C \sslash 2$\\
\hline
\multirow{2}{*}{maxpool} & $4\times7^2$ & \multirow{2}{*}{$C \sslash 2$} \\
& stride $4\times7^2$ &  \\
\hline
conv & $7\times7^2$ & $C$ \\
\hline
globalpool &  & $C$ \\
\hline
fc &  & variable \\
\end{tabular}
\vspace{-2mm}
\caption{Deformation network architecture used in our experiments. The filter sizes are $T\times S^2$ where $T$ is the temporal kernel size and $S$ is the spatial size. The global pool layer averages features across all spatial-temporal locations. ``fc'' is the final fully-connected layer.} 
\label{tab:deformnet}
\vspace{-9mm}
\end{table}
All convolution layers are followed by batch normalization~\cite{ioffe2015batch} and ReLU~\cite{hahnloser2000digital,glorot2011deep}. The dimension of the final FC layer depends on the type of parameterization we choose for the spatial-temporal alignment.
Taking affine transformation for example, the dimension of the network output ($\mathbf{p}_{\text{affine}}=[p_1 \; ... \;p_{12}]^{T}$) is of size 12 and $\theta$ is constructed as 
\begin{align}\label{eqn:affine}
    \theta(\mathbf{p}_{\text{affine}}) = 
\begin{pmatrix}
1 + p_1 & p_2 & p_3 & p_4\\
p_5 & 1 + p_6 & p_7 & p_8\\
p_9 & p_{10} & 1 + p_{11} & p_{12}\\
0 & 0 & 0 & 1\\
\end{pmatrix}
\end{align}
$\theta$ can be more restrictive as in the case of attention~\cite{xu2015show}, where cropping, translation and scaling are allowed for transformation, the dimension of the network output $\mathbf{p}_{\text{att}}=[p_1,...p_6]^T$ is a vector of size 6 and $\theta$ is constructed as 
\begin{align}\label{eqn:att}
    \theta(\mathbf{p}_{\text{att}}) = 
\begin{pmatrix}
1 + p_1 & 0 & 0 & p_4\\
0 & 1 + p_2 & 0 & p_5\\
0 & 0 & 1 + p_3 & p_6\\
0 & 0 & 0 & 1\\
\end{pmatrix}
\end{align}
The design of {\fancyname} is flexible and can be any type of transformation.
The key intuition of the deformation on CNN feature maps is to compensate for the fact that CNNs are not rotation, scale, and shear transformation equivariant~\cite{kosiorek2019stacked}.

\subsection{Warping Module}
\begin{figure}[ht]
	\centering
		\includegraphics[width=0.47\textwidth]{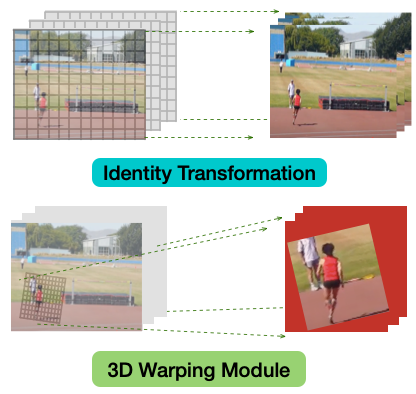} 
	\caption{Two examples of applying spatial-temporal warping/transform to an input sequence. The top row shows an identity transformation while the bottom row shows an example 3D warping where the actor is rotated and zoomed in.
	}
	\label{fig:warp}
\end{figure}

After computing the transformation matrix $\theta$, we utilize a differentiable warping function $\mathcal{T}$ to transform the feature maps with better alignment of the content.
See Fig.~\ref{fig:warp}.
The warping function is essentially a resampling of features from the input feature maps to the output at each corresponding pixel location.
Note that the feature maps could also be images.
Extending from the notation of 2D alignment~\cite{jaderberg2015spatial} , we define the output feature maps $\mathcal{I} \in \mathbf{R}^{C \times T \times H \times W}$ to lie on a spatial-temporal regular grid $G = \{G_i\}$, where each element of the grid $G_i = (t^o_i, x^o_i, y^o_i)$ corresponds to a vector of output features of size $\mathbf{R}^{C}$.  
Hence, the pointwise sampling between the input and output feature maps is written as
\begin{align}
    \begin{bmatrix}
    t^s_i \\
    x^s_i \\
    y^s_i \\
    1 \\
    \end{bmatrix}
    = \mathcal{T}_{g}(\theta, G_i) =
    \begin{pmatrix}
    1\!+\!p_1 & 0 & 0 & p_4\\
    0 & 1\!+\!p_2 & 0 & p_5\\
    0 & 0 & 1\!+\!p_3 & p_6\\
    0 & 0 & 0 & 1\\
    \end{pmatrix}
    \begin{bmatrix}
    t^o_i \\
    x^o_i \\
    y^o_i \\
    1 \\
    \end{bmatrix}
\end{align}
where ($t^o_i, x^o_i, y^o_i$) are the output feature map coordinates in the regular grid and ($t^s_i, x^s_i, y^s_i$) are the corresponding input feature map coordinates for feature sampling.
Here we use attention transformation as an example, where the deformation matrix is $\theta$ parameterized by  $\mathbf{p}_{\text{att}}$ (Eqn.~\ref{eqn:att}).
Given the coordinates mapping, as the computed corresponding coordinates in the input feature maps might not be integers, we utilize the differentiable trilinear interpolation to sample input features from the eight closest points based on their distance to the computed point ($t^s_i, x^s_i, y^s_i$).
In this way, we iterate through every point in the regular grid, ($t^o_i \in [1, ..., T], x^o_i \in [1, ..., W], y^o_i \in [1, ..., H]$) and compute the output feature maps identically for each channel.

\subsection{Channel Grouping}\label{sec:cg}
To increase the flexibility of spatial-temporal alignment, we introduce channel grouping to allow multiple alignments and deformations of feature maps at the same time. 
This is inspired by the fact that there could be multiple attention regions in the video that are relevant to recognizing the actions.
Specifically, we define {\fancyname} with channel grouping as follows:
\begin{align}
\label{eqn:cg}
    \text{STAN}_{cg}(\mathcal{I}_{in}) = \text{concat}\left( \text{STAN}(\mathcal{I}^1_{in}),..., \text{STAN}(\mathcal{I}^{C_g}_{in}) \right)
\end{align}
where ${C_g}$ is the number of channel groups.
Essentially, we apply transformation on each channel group by $\text{STAN}(\mathcal{I}^{C_g}_{in}) \in \mathbf{R}^{C\sslash{C_g} \times T \times H \times W}$ so that multiple alignments can be utilized.

\subsection{Temporal Grouping}\label{sec:tg}
Many actions are composed of a sequence of sub-actions that require different alignments. For example, action ``High Jump'' may consist of ``stand'', ``run'' and ``jump'' (see Fig.~\ref{fig:overview}), and the actor may have moved around between video frames.
Based on this observation, we propose temporal grouping to allow the model to learn different alignments at different time periods.
Similar to channel grouping, it is written as 
\begin{align}
\label{eqn:cg}
    \text{STAN}_{tg}(\mathcal{I}_{in}) = \text{concat}\left(\text{STAN}(\mathcal{I}^1_{in}),..., \text{STAN}(\mathcal{I}^{T_g}_{in}) \right)
\end{align}
where ${T_g}$ is the number of temporal groups and concatenation is operated on the temporal axis.
Each group alignment $\text{STAN}(\mathcal{I}^{T_g}_{in})$ is of size $\mathbf{R}^{C \times T\sslash{T_g} \times H \times W}$.

\subsection{Integration with Backbone CNNs}\label{sec:arch}
Now that we have defined {\fancyname} function we now discuss effective ways to add it into existing backbone CNN networks. In this paper, we focus on designing {\fancyname} for the RGB stream and the extension to optical flow will be left as one of our future work.
Recent works~\cite{wang2018non,feichtenhofer2019slowfast} and their variants~\cite{pan2020actor} achieve single-stream state-of-the-art performance on action recognition and detection with 3D CNNs. 
We therefore explore adding {\fancyname} layers to the ResNet3D model~\cite{wang2018non,feichtenhofer2019slowfast} and the SlowFast~\cite{feichtenhofer2019slowfast} network.
Intuitively, placing {\fancyname} layer at shallower layers allows earlier feature alignments that could potentially lead to cleaner representation for better action recognition. However, shallow layers may not have enough abstraction in the feature maps for {\fancyname} to learn the right alignments.
Based on this trade-off, we experiment with adding one {\fancyname} layer after $res_2$ block. 
For SlowFast, we only add {\fancyname} layers on the Fast pathway since it contains more temporal information compared to the Slow pathway.
Other insertion locations are explored in the ablation experiments (Section~\ref{sec:ablation}).

\subsection{Detection Architecture}\label{sec:detect}
For action recognition, the final CNN outputs are passed through global averaging (and a concatenation for SlowFast) and a fully-connected layer to get the action class probabilities.
For action detection, following previous works~\cite{gu2018ava,sun2018actor,feichtenhofer2019slowfast}, we use pre-computed actor proposals.
The bounding boxes are used to extract region-of-interest (RoI)
features using RoI-Align~\cite{he2017mask} at the last feature map of ``$res_5$'' after temporal global average pooling. 
Our {\fancyname} can fix the actor misalignment problem by aligning the 2D RoIs along the temporal axis. 
Finally, the RoI features are then max-pooled and fed to a fully-connected layer.

\section{Experiments}
\label{sec:exp}
To demonstrate the efficacy of our models, specifically, the viewpoint-invariant design that helps action models to generalize, we experiment on two recent action detection datasets, including AVA~\cite{gu2018ava} and AVA-Kinetics~\cite{li2020ava}, and several major action recognition datasets, Kinetics-400~\cite{kay2017kinetics}, Charades~\cite{sigurdsson2016hollywood}, Charades-Ego~\cite{sigurdsson2018charades}.

\subsection{Action Detection}
\label{sec:exp_detect}
This section evaluates our {\fancyname} model for the task of spatial temporal localization for actions using AVA and AVA-Kinetics.
We show significant improvement over baselines while only adding a fraction of computation. 

\noindent\textbf{Datasets.}
The AVA dataset~\cite{gu2018ava} is an action detection dataset where models are required to output action classification results with bounding boxes. 
Spatial-temporal labels are provided for one frame per second, with every person annotated with a bounding box and (possibly multiple) actions.
There are 211k training and 57k validation video clips. 
We use AVA v2.1 and follow the standard protocol~\cite{gu2018ava,feichtenhofer2019slowfast} of evaluating on 60 classes.
The performance metric is mean Average Precision (mAP) over 60 classes, using an IoU threshold of 0.5.
We report mAP on the official validation set.

The AVA-Kinetics dataset~\cite{li2020ava} is a recent action detection dataset which follows AVA annotation protocol to annotate relevant videos on the Kinetics-700~\cite{carreira2019short} dataset.
We utilize the part where all videos are from Kinetics for training and testing.
There are about 141k training videos and 32k validation videos.
The AVA-Kinetics also contains the same 60 classes for evaluation and we utilize the mAP metric. Following~\cite{li2020ava}, we evaluate action detection model performance with both ground truth person boxes and pre-computed person boxes.
We report mAP on the official validation set as well.

\noindent\textbf{Action Proposals.}
The action detection models have to output action/person bounding boxes, and only predicted boxes with IoU (intersection over union) area w.r.t the ground truth boxes above a threshold of 0.5 are considered true positives.
As mentioned in Section~\ref{sec:detect}, we follow previous works~\cite{gu2018ava,sun2018actor,feichtenhofer2019slowfast} and use pre-computed person boxes as action proposals.
For AVA, we utilize the same person proposals from ~\cite{feichtenhofer2019slowfast}.
For AVA-Kinetics, we finetune a Mask-RCNN~\cite{he2017mask} with ResNet-101 backbone trained on COCO~\cite{lin2014microsoft} on the person boxes in the training set and extract boxes as described in Section~\ref{sec:detect}.
The region proposals for action detection are detected person boxes with a confidence score of larger than 0.8, which has a recall of 83.1\% and a precision of 62.1\% for the person class, given IoU threshold of 0.5.
The average precision of the person class is 0.732 on the validation set.

\noindent\textbf{Training.}
We initialize the network weights from the Kinetics-400 classification models, following previous works~\cite{feichtenhofer2019slowfast}. 
We use a learning rate of 0.1 and cosine learning rate decay. We use synchronized SGD to train for 10 epochs with a batch size of 16 on a 4-GPU machine, with a linear warm-up from 0.000125 for the first 2 epochs.
We use a weight decay of $10^{-7}$. 
Ground-truth boxes and video clips centered at the annotated key frames are sampled for training. The video is first resized to 256x320, and then 
we use random 224×224 crops and horizontal flipping  following~\cite{feichtenhofer2019slowfast}.

\noindent\textbf{Inference.}
Since the annotations on AVA and AVA-Kinetics are one (key) frame per second, we sample a single video clip temporally centered around the key frame for evaluation.
Following~\cite{feichtenhofer2019slowfast}, we resize the spatial dimension such that its shorter side is 256 pixels. 
Ground truth boxes or pre-computed boxes are used as inputs.
We report the inference time computational cost (in FLOPs) of a single 256x320 clip using Tensorflow's Profiler. 

\noindent\textbf{Implementation Details.}
We add one {\fancyname} layer to the backbone CNN network as described in Section~\ref{sec:arch}.
We use affine transformation (Eqn.~\ref{eqn:affine}) and the deformation network is defined in Table~\ref{tab:deformnet}.
We use a temporal group of 2 and the number of base convolution filters in the deformation network is capped at 8 for ResNet3D to keep the FLOPs low. For SlowFast, we set this number as described in Table~\ref{tab:deformnet}.

\noindent\textbf{Baselines.}
To demonstrate the effectiveness of our proposed model, we experiment with recent 3D-CNN based models for action recognition and detection.
\textbf{ResNet3D} is a model based on ResNet-50~\cite{he2016deep} with additional 3D convolutional filters. 
\eat{
The architecture is the same as the SLOWONLY model in ~\cite{feichtenhofer2019slowfast}.
}
The number of input frames is 8 and we sample 1 frame every 8 frames (i.e., 8x8 frames).
\textbf{ResNet3D + {\fancyname}} is our proposed model added to ResNet3D as described in Section~\ref{sec:arch} with the same inputs.
\textbf{SlowFast} is a recent efficient model~\cite{feichtenhofer2019slowfast} with a Slow pathway and a Fast pathway, which takes 8x8 and 32x2 frames as inputs respectively. We use ResNet-50 backbone for SlowFast as well.
If we use only the Slow pathway, the model will become the same as ResNet3D. 
\textbf{SlowFast + {\fancyname}} is our proposed model added to SlowFast as described in Section~\ref{sec:arch}.

\begin{table}[t!]
\centering
\begin{tabular}{l|c|c|c}

Models                & mAP  & GFLOPs & MParams  \\ \hline
ResNet3D   (8x8)            & 0.234 &   208.0 & 31.75      \\ 
ResNet3D + {\fancyname} & \textbf{0.241} & 216.6 & 32.02  \\ \hline
SlowFast~\cite{feichtenhofer2019slowfast}    (32x2)              & 0.252&     242.6 & 33.77     \\ 
SlowFast + {\fancyname} & \textbf{0.263} & 247.4 & 33.96 \\ 
\end{tabular}
\caption{Experiment results on AVA dataset. We show the model performance on mAP, computation cost (in billions) and number of parameters (in millions). The computational cost is of a single 256x320 video clip of length 8x8 or 32x2 (number of frames x sampling rate) frames.
}
\label{tab:ava}
\end{table}

\begin{table}[t!]
\centering
\begin{tabular}{l|c|c}

Models                & mAP & GFLOPs \\ \hline
Action Transformer~\cite{li2020ava}    & 0.337 / 0.191  & -                \\ \hline
ResNet3D  (8x8)            & 0.315 / 0.224      & 208.0                 \\ 
ResNet3D + {\fancyname} & \textbf{0.336} / \textbf{0.238}  & 216.6      \\ \hline
SlowFast~\cite{feichtenhofer2019slowfast}         (32x2)           & 0.341 / 0.242 &     242.6                \\ 
SlowFast + {\fancyname} & \textbf{0.358} / \textbf{0.253}  & 247.4  \\ 
\end{tabular}
\caption{Experiment results on AVA-Kinetics dataset. We show both mAP with ground truth boxes and detected boxes. We are not able to get Action Transformer's FLOPs as the code is not public. Note that the pre-computed boxes for Action Transformer is different from ours.}
\label{tab:ava-kinetics}
\end{table}

\subsubsection{Main Results}
\noindent\textbf{AVA.}
Table~\ref{tab:ava} shows the results on AVA dataset~\cite{gu2018ava}.
We follow the SlowFast~\cite{feichtenhofer2019slowfast} paper's evaluation protocol and use the same predicted person bounding boxes provided by the authors with ROIAlign to classify actions.
For both ResNet3D and SlowFast, we use ResNet-50 as their base architecture.
Compared with previous methods, our model is able to achieve 3\% relative improvement on mAP for ResNet3D, with only relatively 4\% more computation.
The parameter increase is also minimal.
On SlowFast, our method's improvement is more efficient, with 4\% \textbf{relative} improvement
on mAP and only 2\% more computation.

\noindent\textbf{AVA-Kinetics} is a recent action detection dataset with Internet videos from the Kinetics-700 dataset~\cite{carreira2019short}.
Table~\ref{tab:ava-kinetics} shows the results on AVA-Kinetics dataset~\cite{li2020ava}.
We follow the baseline~\cite{li2020ava} paper's evaluation protocol and experiment with both ground truth person boxes and detected person boxes from the described finetuned Mask-RCNN model.
Our model is able to achieve 7\% relative improvement on mAP for ResNet3D with only relatively 4\% more computation and 5\% improvement with 2\% more computation on SlowFast.

\subsubsection{Ablation Experiments}\label{sec:ablation}

\begin{table}[t!]
\centering
\begin{tabular}{l|c|c|c}

               & Diff & mAP & GFLOPs   \\ \hline
SlowFast             & - & 0.252 &   242.55     \\ 
+ {\fancyname} & \textbf{+1.1\%} & \textbf{0.263} & 247.40 \\ \hline
+ {\fancyname} ($res_3$) & +0.7\% & 0.259 & 247.35 \\
+ {\fancyname} ($res_4$) & +0.3\% & 0.255 & 247.39 \\ \hline
+ {\fancyname} (Att, 6) & +0.7\% & 0.259 & 247.40  \\
+ {\fancyname} (Sp, 6) & +0.6\% & 0.258 & 247.40  \\
+ {\fancyname} (H, 15) &  +0.3\% & 0.255 & 247.40  \\ \hline
+ {\fancyname} (no tg) &  +0.8\% & 0.260 & 247.40  \\
+ {\fancyname} (cg=8) &  +1.0\% & 0.262 & \textbf{243.30}  \\ \hline
+ {\fancyname} (fixed $W_{\theta}$) &  +0.7\% & 0.259 & 247.40 \\
\end{tabular}
\caption{Ablation experiment results (on AVA). The computational cost is of a single 256x320 video clip of length 32x2 (number of frames x sampling rate) frames. The ``Diff'' column shows absolute improvement of the variant models compared to the baseline SlowFast model. ``(Att, 6)'' means attention transformation with 6 degree-of-freedom (DoF), ``(Sp, 6)'' means spatial transformation with 6 DoF (defined in Eqn.~\ref{eqn:sp}) and ``(H, 15)'' means homography transformation with 15 DoF. ``tg'' and ``cg'' means temporal grouping and channel grouping, respectively. See text for details.}
\label{tab:ablation}
\end{table}

In this section, we perform ablation studies on the AVA dataset with the SlowFast model as the backbone network.
To understand how action models can benefit from {\fancyname}, we explore the following questions (results are shown in Table~\ref{tab:ablation}):

\noindent\textbf{Where to insert {\fancyname} layer?}
In CNN networks, shallower layers tend to encode low-level visual features like edges and patterns while deeper layers may contain more abstract information.
Placing {\fancyname} layer at shallower layers allows earlier feature alignments that could potentially lead to cleaner representation. 
To verify this hypothesis, we experiment with adding one {\fancyname} layer at deeper layers than $res_2$ block. 
We add {\fancyname} layer after $res_3$ and $res_4$.
As we see, the model performance deteriorates significantly, suggesting that {\fancyname} should be placed at earlier layers.

\noindent\textbf{What is the best parameterization for the deformation network?}
In Section~\ref{sec:deformnet}, we have discussed two ways of parameterization for the deformation network, affine transformation (Eqn.~\ref{eqn:affine}) and attention transformation (Eqn.~\ref{eqn:att}). 
Each has 12 and 6 degree-of-freedom (DoF), respectively.
We use affine transformation in our main experiments and here we experiment with attention transformation, 
spatial transformation
and homography transformation.
Given the deformation network output of $\mathbf{p}_{\text{sp}}=[p_1,...p_6]^T$, the spatial transformation is defined as follows:
\begin{align}\label{eqn:sp}
    \theta(\mathbf{p}_{\text{sp}}) = 
\begin{pmatrix}
1 + p_3 & 0 & 0 & p_4\\
0 & 1 + p_1 & -p_2 & p_5\\
0 & p_2 & 1 + p_1 & p_6\\
0 & 0 & 0 & 1\\
\end{pmatrix}
\end{align}
where the second and third row is for the width and height dimension.
For homography transformation, the last element in the $4\times4$ matrix is set as 1 hence the DoF is 15, which is shown in Table~\ref{tab:ablation} (``H, 15'').
Results are shown in Table~\ref{tab:ablation}. As we see, the model with the most free parameters performs the worst, while attention and spatial transformation perform worse than affine transformation.

\noindent\textbf{Does channel/temporal grouping help?}
In this experiment, we validate the efficacy of temporal grouping (Section~\ref{sec:tg}) and channel grouping (Section~\ref{sec:cg}). 
The main experiments are conducted with a temporal group of 2, and the model performance drops by a small margin if temporal grouping is removed.
Interestingly, when we apply a channel grouping of 8 to the original model, we can achieve similar performance improvement but with significantly less computation (with channel grouping, we only need to add 0.3\% relatively more computation compared to 2\% as before to achieve nearly the same performance.)
This result is also observed in the Kinetics-400 experiments (Table~\ref{tab:kinetics}).

\noindent\textbf{Does {\fancyname} transfer well?}
Finally, we conduct an experiment to see whether the deformation network learned from a dataset can be generalized to another.
We train the original {\fancyname} model on Kinetics-400 dataset, and then only fine-tune the layers after the {\fancyname} layer on AVA.
As we see in Table~\ref{tab:ablation} (``fixed $W_{\theta}$''), {\fancyname} can still achieve reasonable improvement on AVA, suggesting the deformation can be transferred from one dataset to another.

\begin{table}[]
\centering
\begin{tabular}{l|c|c|c}

        Models              & top-1          & top-5   & GFLOPs       \\ \hline
NonLocal R50~\cite{wang2018non}     & 0.749          & 0.916 & -          \\ \hline
ResNet3D (8x8)             & 0.735          & 0.908   & 109.2       \\ 
ResNet3D + {\fancyname} & \textbf{0.751} & \textbf{0.916} & 113.2 \\ \hline
SlowFast~\cite{feichtenhofer2019slowfast}  (32x2)           & 0.759          & 0.920     & 131.7     \\
SlowFast + {\fancyname} & \textbf{0.769} & \textbf{0.928} & 134.5\\ 
SlowFast + {\fancyname} (cg=8) & \textbf{0.770} & \textbf{0.928} & 132.3\\ 
\end{tabular}
\caption{Experiment results on Kinetics-400 dataset. The nonlocal baseline has an input of 32 frames. ``cg'' means channel grouping. The computational cost is of a single 256x256 video clip. }
\label{tab:kinetics}
\end{table}

\subsection{Action Recognition}
\label{sec:exp_rec}
The action recognition task is defined to be a classification task given a trimmed video clip. To evaluate the generalization abilities of our proposed model, we consider three major datasets, Kinetics-400~\cite{kay2017kinetics}, Charades~\cite{sigurdsson2016hollywood} and Charades-Ego~\cite{sigurdsson2018charades}.

\noindent\textbf{Datasets.}
Kinetics-400~\cite{kay2017kinetics} consists of about 240k training videos and 20k validation videos in 400 human action classes. The videos are about 10 seconds long. Following previous works~\cite{feichtenhofer2019slowfast,wang2018non}, we report top-1 and top-5 classification accuracy.
Charades~\cite{sigurdsson2016hollywood} is a dataset with longer ( about 30 seconds on average) videos of indoor activities. There are about 9.8k training videos and 1.8k validation videos in 157 classes in a multi-label classification setting.
Charades-Ego~\cite{sigurdsson2018charades} has the same 157 action labels but consists of both third-person view and first-person view videos. Essentially, this dataset shows different perspectives of the same actions, which makes it ideal to test our proposed method.
Performance is measured in mean Average Precision (mAP).

\noindent\textbf{Training.}
Our models on Kinetics are trained from scratch with random initialization, without using any pre-training (same as in ~\cite{feichtenhofer2019slowfast}). 
We use a learning rate of 0.2 and cosine learning rate decay. We use synchronized SGD to train for 100 epochs with a batch size of 16 on a 4-GPU machine, with a linear warm-up from 0.01 for the first 20 epochs.
We use a weight decay of $10^{-7}$. 
On Charades and Charades-Ego, we initialize the models using models trained on Kinetics-400.
We use a learning rate of 0.02 and train for 50 epochs, where learning rate reduces to its 1/10 at epoch 40. We use a linear warm-up from 0.000125 for the first 2 epochs.
For all three datasets, we use random $224\times224$ crops and horizontal flipping from a video clip, which is randomly sampled from the full-length video and resized to a shorter edge side of randomly sampled in [256, 320] pixels.

\noindent\textbf{Inference.}
Following previous works~\cite{feichtenhofer2019slowfast,wang2018non}, we sample $3\times10$ clips for each video during testing: we uniformly sample 10 clips for the temporal domain and 3 spatial crops of size $256\times256$ after the shorter edge size are resized to 256 pixels.
We average the softmax scores across all clips for final prediction for Kinetics-400 and use the maximum of the softmax scores for Charades and Charades-Ego.
We report the computational cost of a single, spatially center-cropped clip of size $256\times256$.

\subsubsection{Recognition Results}

We compare the same baselines, as mentioned in the previous section. 
For Kinetics-400, the input frames are the same as the previous section. For Charades and Charades-Ego, we use $16\times8$ input frames for ResNet3D and $32\times4$ input frames for SlowFast.

\noindent\textbf{Kinetics-400.} Table~\ref{tab:kinetics} shows the experiments on Kinetics-400. Our method can improve top-1 accuracy by 1.6 and 1 point for ResNet3D and SlowFast, respectively, at the cost of 3.6\% and 2.1\% relatively more computation.
In addition, we experiment with channel grouping with 8 groups (cg=8). The improvement of accuracy is similar, but due to a smaller number of filters per group, the computation cost is further reduced.

\begin{table}[t!]
\centering
\begin{tabular}{l|c|c|c}
Models         & mAP  & GFLOPs & MParams\\ \hline
ResNet3D  (16x8)            &  0.354   &    218.4 & 32.40   \\ 
ResNet3D + {\fancyname} & \textbf{0.377} & 226.4 & 32.47 \\ \hline
SlowFast~\cite{feichtenhofer2019slowfast}   (32x4)       & 0.386 & 131.7 & 34.51        \\ 
SlowFast + {\fancyname} & \textbf{0.406}  & 134.5 & 34.53    \\
\end{tabular}
\caption{Experiment results on the Charades dataset. We show the model performance on mAP, computation cost (in billions), and number of parameters (in millions). The computational cost is of a single 256x256 video clip of length 16x8 or 32x4 (number of frames x sampling rate) frames.}
\label{tab:charades}
\end{table}

\noindent\textbf{Charades.} Table~\ref{tab:charades} shows the experiments on the Charades dataset. 
Our method is able to improve mAP by 2.3 and 2 points for ResNet3D and SlowFast, respectively, with only 3.6\% and 2.1\% relatively more computation.
Our method only contains very minimal parameters.

\begin{table}[t!]
\centering
\begin{tabular}{l|c|c}
Models         & 1st-person & 3rd-person  \\ \hline
Baseline v1.0~\cite{sigurdsson2018charades}   &    0.282                           & 0.232     \\ \hline
ResNet3D  (16x8)            & 0.298                           & 0.361   \\ 
ResNet3D + {\fancyname} & \textbf{0.318}                  & \textbf{0.366} \\ \hline
SlowFast~\cite{feichtenhofer2019slowfast}   (32x4)      & 0.316                           & 0.391    \\ 
SlowFast + {\fancyname} & \textbf{0.326}                  & \textbf{0.396}   \\
\end{tabular}
\caption{Experiment results on Charades-Ego dataset. Models are trained on both 1st-person and 3rd-person view videos. We show the test set with 1st-person and 3rd-person view videos separately. The computation cost and number of parameters are the same as the Charades experiment.}
\label{tab:charades-ego}
\end{table}

\noindent\textbf{Charades-Ego.} Table~\ref{tab:charades-ego} show the results on the Charades-Ego dataset.
The test set is divided into 1st-person videos and 3rd-person videos. Note that the training set of Charades-Ego dataset includes mostly 3rd-person videos, and we can observe from the results that our {\fancyname} model achieves more significant improvement on the 1st-person test set, verifying the efficacy of our model's generalization ability.
With SlowFast and {\fancyname} model, we are able to achieve state-of-the-art performance on this dataset.

\begin{figure}[t!]
	\centering
		\includegraphics[width=0.47\textwidth]{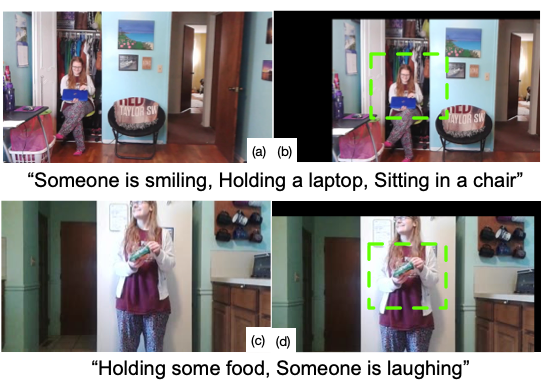} 
	\caption{Qualitative analysis. Videos are from Charades. (a) and (c) are original frames. (b) and (d) are  transformed frames. The green bounding box denotes a center reference. The sentences below the images are the ground truth labels for the video clip. See main text for details of analysis.
	}
	\label{fig:qual}
\end{figure}

\subsubsection{Qualitative Analysis.}
{\fancyname} can provide intuitive geometric interpretation of human actions.  
To illustrate such effects, we visualize example transformations of our model on Charades videos on two videos with correct classification labels in Fig.~\ref{fig:qual}. We use the output ($\theta$, Eqn.~\ref{eqn:affine}) from the {\fancyname} layer (deformation network) of the first temporal group and visualize the spatial transformation using the middle frame. 
Fig.~\ref{fig:qual} (a) and (c) are the original frames.
Fig.~\ref{fig:qual} (b) and (d) are the transformed frames using the transformation matrix predicted by {\fancyname}.
We use a center bounding box as a reference in the visualization. As we see, for the first example, the person originally is located to the left of the scene and {\fancyname} learns a transformation that centers the main actor.
In the second example, the person is already in the center and the transformation does not alter much of the frames.

\section{Conclusion}
\label{sec:concl}
This paper has introduced a new spatial-temporal alignment network, {\fancyname}, for action recognition and detection.
Our study is the first to explore explicit spatial-temporal alignment in 3D CNNs for action detection.
Our model can be conveniently inserted into existing networks and provides significant improvement with a low extra computation cost.
We have shown that our method achieves state-of-the-art performance on multiple challenging action recognition and detection benchmarks.
We believe our models will facilitate future research and applications on viewpoint-invariant feature representation learning for actions. In the future, we would like to extend this work to the optical flow stream and investigate how to learn the two-stream network more accurately and more efficiently. 

{
\bibliographystyle{ieee_fullname}
\bibliography{egbib}
}
\end{document}